# Utilizing AI for Aviation Post-Accident Analysis Classification

Aziida Nanyonga[1] and Graham Wild[2]
*UNSW, Canberra, ACT, 2612, Australia*

The volume of textual data available in aviation safety reports presents a challenge for timely and accurate analysis. This paper examines how Artificial Intelligence (AI) and, specifically, Natural Language Processing (NLP) can automate the process of extracting valuable insights from this data, ultimately enhancing aviation safety. The paper reviews ongoing efforts focused on the application of NLP and deep learning to aviation safety reports, with the goal of classifying the level of damage to an aircraft and identifying the phase of flight during which safety occurrences happen. Additionally, the paper explores the use of Topic Modeling (TM) to uncover latent thematic structures within aviation incident reports, aiming to identify recurring patterns and potential areas for safety improvement. The paper compares and contrasts the performance of various deep learning models and TM techniques applied to datasets from the National Transportation Safety Board (NTSB) and the Australian Transport Safety Bureau (ATSB), as well as the Aviation Safety Network (ASN), discussing the impact of dataset size and source on the accuracy of the analysis. The findings demonstrate that both NLP and deep learning, as well as TM, can significantly improve the efficiency and accuracy of aviation safety analysis, paving the way for more proactive safety management and risk mitigation strategies.

## I.  Nomenclature

| | | |
|---|---|---|
| ASN     | = | Aviation Safety Network |
| ASRS    | = | Aviation Safety Reporting System |
| ATSB    | = | Australian Transport Safety Bureau |
| BLSTM   | = | Bidirectional Long Short-Term Memory |
| CNN     | = | Convolutional Neural Networks |
| GRU     | = | Gated Recurrent Units |
| IASMS   | = | In-time Aviation Safety Management System |
| ISSA    | = | In-Time System-wide Safety Assurance |
| LDA     | = | Latent Dirichlet Allocation |
| LSA     | = | Latent Semantic Analysis |
| ML      | = | Machine Learning |
| NLP     | = | Natural Language Processing |
| NMF     | = | Non-negative Matrix Factorization |
| NTSB    | = | National Transport Safety Board |
| SMS     | = | safety management system |
| LSTM    | = | Long Short-Term Memory |
| pLSA    | = | Probabilistic Latent Semantic Analysis |
| ResNet  | = | Residual Networks |
| RNN     | = | Recurrent Neural Networks |
| sRNN    | = | Simple Recurrent Neural Network |
| TM      | = | Topic Modelling |

---

[1] PhD Candidate, School of Engineering and Technology.
[2] Senior Lecturer, School of Science, and AIAA Member.

## II. Introduction

It is anticipated that the future aerospace transportation system will involve a significant increase in traffic, both conventional and urban based [1]. There are a number of barriers to the safe integration to future technologies like urban air mobility [2], with automation being a key aspect. Data is also another key aspect, with ISSA intended to "mitigates risks before they can lead to an incident or accident" [3]. The ISSA is part of the designed IASMS. Investigation is an important aspect of safety assurance and SMS in general [4]. To take advantage of automation and the "big data" available, the use of artificial intelligence and machine learning is essential. Starting with more general applications in aviation and aerospace, such as onboard fault monitoring and diagnosis [5], intelligent decision support [6], as well as planning and operations [7], the growing application to safety is key [8-11].

Safety is critical, with the aviation industry prioritizing safety to ensure passenger and crew well-being and to maintain public trust in air travel [12]. To achieve this, post-accident analysis is essential in identifying the causes of aviation incidents and preventing future occurrences [13]. Traditionally, safety analysis has relied heavily on manual inspection and categorization of incident reports, a time-consuming process susceptible to human error [14]. Experts often manually review narratives, findings, and recommendations to find recurring patterns and contributing factors in accident reports; this approach, though valuable, is limited by human capacity, subjectivity, and the potential for oversight [15].

However, the industry generates a large amount of safety data through various reporting systems, such as the ASRS and the NTSB [16]. These reports, often rich in narrative content, provide detailed accounts of incidents and accidents [17]. The unstructured nature of pure text narratives makes it difficult to use traditional data analysis methods, which are typically designed for structured data. Recent advancements in NLP and ML present opportunities to improve the analysis of aviation safety data. These technologies offer efficient and accurate methods to uncover insights from large volumes of text-based reports [18].

This paper will summarize research on the use of AI and ML for post-accident analysis classification. The research explores how these technologies can transform the field by providing more reliable and timely insights, ultimately contributing to improved aviation safety. Specifically, this research examines how AI and ML can be used to:
- Improve the efficiency of safety reporting and analysis processes.
- Enhance the accuracy of safety analysis.
- Identify latent safety issues.
- Gain deeper insights into safety-related incidents and causal factors.
- Enhance predictive capabilities for identifying potential safety risks.

This research suggests that NLP and deep learning can effectively extract information from raw text narratives and facilitate the thorough analysis of safety occurrences in the aviation industry.

## III. Background Research

Initial research examined the use of NLP and deep learning models to classify the level of damage to an aircraft following an aviation safety occurrence, using the textual narrative of pre-accident events from NTSB reports [8]. We trained and evaluated various deep learning models, including LSTM, BLSTM, GRU, sRNN, and combinations thereof, on a dataset of over 27,000 safety occurrence reports. We found that all models achieved high accuracy, exceeding 87.9%, with sRNN and joint RNN-based models like GRU+LSTM and sRNN+BLSTM+GRU performing particularly well. We compared this with another study utilizing the ATSB dataset [19], with a comparative study of the same four deep learning models—sRNN, LSTM, BLSTM, and GRU—for classifying aircraft damage levels based on textual descriptions from aviation safety reports. The analysis showed that all models achieved high accuracy (over 88%), significantly exceeding a random guess, with the sRNN model demonstrating superior performance in terms of recall and overall accuracy (89%).

Next, we explored using NLP and deep learning (ResNet and sRNN models) to analyze aviation safety reports from the NTSB [20]. The aim was to automatically classify the phase of flight (e.g., takeoff, landing) during which safety occurrences happened, using the unstructured text narratives within those reports. The analysis of a dataset of 27,000 reports showed that both models achieved accuracy exceeding 68%, with sRNN outperforming ResNet, suggesting a viable method for improving aviation safety analysis and investigation efficiency. Further work was applied to identify the phase of flight associated with an aviation safety incident based on the textual narratives found in incident reports [21, 22]. We examined four deep learning models: LSTM, CNN, BLSTM, and sRNN, using a dataset of over 50,000 safety reports from the ATSB, achieving an accuracy of around 87% with the LSTM and BLSTM models [21]. We also investigated LSTM, GRU, and BLSTM models, as well as combinations of those models, training on a dataset of over 4,000 reports from the ASN and achieving the best results with the LSTM+BLSTM model at 67% accuracy [22].

In addition to wanting to look at performance for generating outputs, we also wanted to understand how performance was affected by inputs. As such, we investigated how the size and source of datasets affect the accuracy of machine learning models used to predict aviation safety incidents [23]. Again, we used NLP techniques to analyze textual data from two datasets: the ASN and the NTSB, employing four neural network architectures (LSTM, GRU, CNN, and RNN). The key finding is that larger datasets consistently improve predictive accuracy, with the NTSB dataset, despite having the same number of initial instances, outperforming ASN. CNNs demonstrated the best performance across various metrics. The study highlights the importance of adequate data collection for improving aviation safety prediction.

Next, we wanted to investigate TM by comparing two techniques, LDA and NMF, for analyzing aviation accident reports from the NTSB dataset [15]. The core goal is to automate the identification of recurring themes and patterns within these reports to improve aviation safety. The study uses topic modeling to extract latent topics, evaluating model performance using topic coherence ($C\_v$). LDA demonstrated superior topic coherence, suggesting stronger semantic relationships within extracted themes, while NMF produced more granular and distinct topics, useful for focused analysis of specific accident aspects. Next we applied four TM techniques, pLSA, LSA, LDA, and NMF, to analyze textual data from ATSB aviation accident reports [24]. As before, we use NLP to preprocess the data before applying these techniques, aiming to automatically identify underlying themes and patterns within the reports. Finally, we have applied TM to narratives from the NTSB dataset, using all four techniques, LDA, NMF, LSA, and pLSA, along with K-means clustering, to identify latent themes, explore semantic relationships, and group similar incidents [25]. The comparative analysis reveals LDA as the most effective technique, and the results uncover recurring themes such as mechanical failures and fuel system issues.

## IV. Comparative Analysis

### A. Damage Level Identification

First, let's consider the damage level, comparing performance as a function of data source [8, 19]. In this both studies utilized NLP and deep learning models to analyze textual narratives of aviation incidents and classify the level of damage to the aircraft. The underlying aim was to improve upon traditional, manual methods of damage level classification, enabling a more efficient and accurate understanding of safety occurrence severity to improve aviation safety. Both studies utilize the same methodology in terms of data preprocessing, model selection, training, and evaluation, enabling a direct side by side comparison. This includes tokenization, encoding categorical data, and addressing special characters and stop words. Similarly, both studies trained the models on a subset of their data and use a separate portion for testing the model's performance. The same four deep learning models: sRNN, LSTM, BLSTM, and GRU; where all of these models are types of RNNs, which are particularly suited for processing sequential data like text.

The key difference between the studies to facilitate a comparison lies in the datasets employed. Our more recent study uses a dataset of 50,778 safety occurrence reports from the ATSB spanning from 2013 to 2022 [19]. The initial study used a dataset of 27,000 safety occurrence reports from the NTSB from 2005 to 2020 [8]. While it is important to note the difference in the date range, the key difference is in the sample size available

While our initial research investigated the four individual models and their combinations [8], the gains in performance for the combinations was not significant enough to continues with their use, given the significant increase in computational time. That is, the newer study only focused on comparing the performance of the four individual deep learning models [19].

Interestingly, the size of the ATSB dataset did not result in an improvement in performance, with both studies finding accuracies of 88%. It should be noted that while the ATSB dataset is larger in terms of the number of safety occurrences, the narrative lengths involved were significantly different to that of the NTSB dataset. That is, two variables were changed simultaneously, the number of text narratives used to train, and the size of the text narratives. Fig. 1 shows the distribution of text narrative sizes for the NTSB, ATSB, and ASN. Here it can be seen that the average length of NTSB reports is greater, with a greater distribution of longer narratives, while the ATSB dataset includes a much greater variation from very short to relatively long text narratives. Specifically, the NTSB had a max narrative length of 2228 words, with an average narrative length of 107.5 words, while the ATSB had a max narrative length of 1264 words, with an average narrative length of 20.4 words. As such, the total corpus for the NTSB was 2.9 million words, while for the ATSB it was 1 million words. Clearly there is an influence on the number and length of narrative utilized. Future research clearly needs to investigate the relative impact of each of these dimensions separately, using different narrative lengths for the same data set, as well as different number of narratives with similar narrative lengths.

### B. Flight Phase Identification

Next is phase of flight classification, comparing performance as a function of data source. All three studies utilized NLP and deep learning models to analyze aviation safety reports and classify the phase of flight during which incidents occurred [20-22]. Again, the underlying goal was to automate this classification process, traditionally done manually, to improve the efficiency and accuracy of safety occurrence analysis for aviation stakeholders.

All three studies share a similar methodology involving data preprocessing, model selection, training, and evaluation. Preprocessing steps include converting text into numerical data, such as tokenization, encoding categorical data, and addressing special characters and stop words. Each study trained their models on a subset of their data and used a separate portion for testing the model's performance.

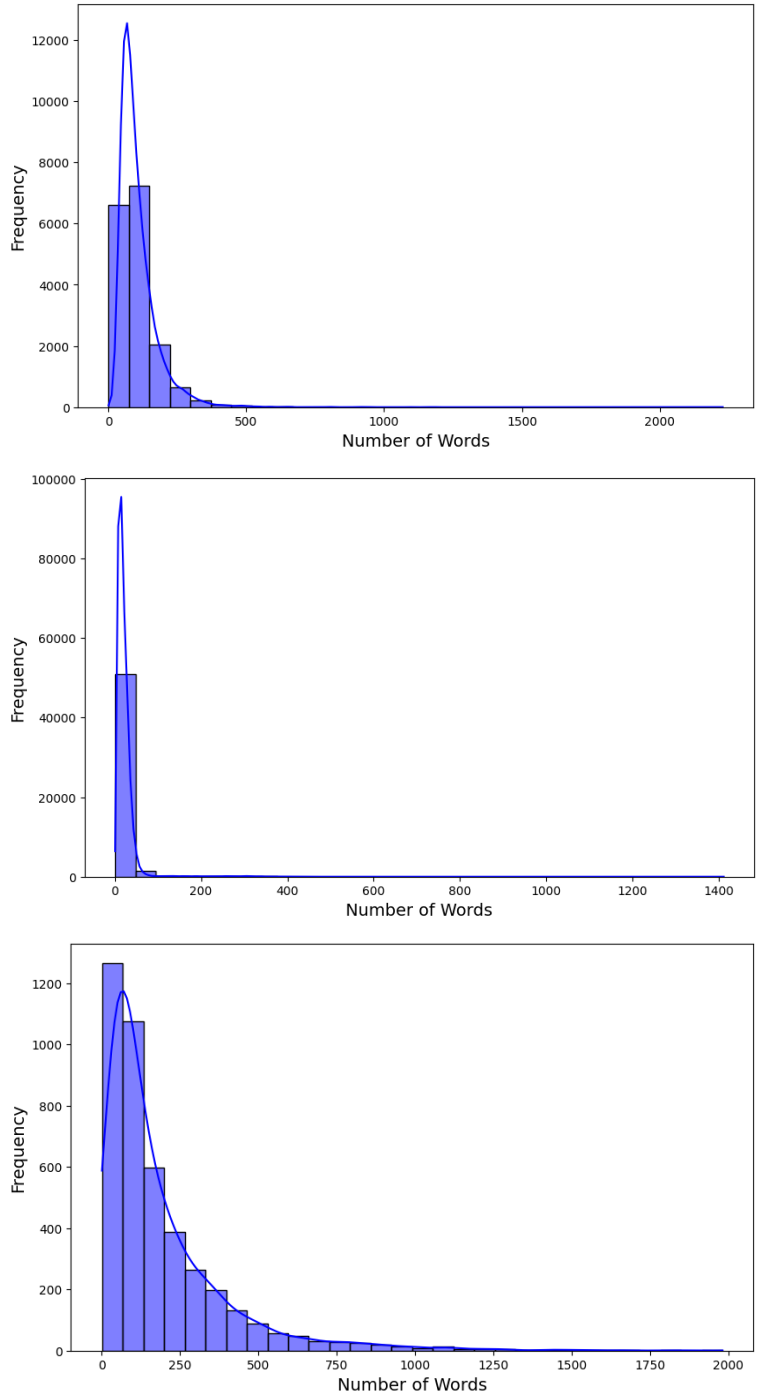

**Fig. 1  Narrative distributions of the three data sources, NTSB (top), ATSB (middle), and ASN (bottom).**

Although there was an overlap in the deep learning models tested, not all studies used the same models. The sRNN model is common to all three papers [20-22]. The LSTM and BLSTM models are used in the [21] and [22]. While [20] used a simplified ResNet model in addition to the sRNN. Also, [22] included a GRU model along with sRNN, LSTM, and BLSTM, and also tests combinations of models like GRU+LSTM, GRU+BLSTM+LSTM, BLSTM+LSTM, and GRU+BLSTM.

The data sets used in the studies were the main difference:
1) [20] utilized 27,000 safety occurrence reports from the NTSB, from 2005 to 2020,
2) [21] utilized 53,275 safety occurrence reports from the ATSB, from 2013 to 2022, and
3) [22] utilized 4,372 aviation investigation reports from the ASN website from 2000 to 2020.

All three studies evaluate model performance using metrics like accuracy, precision, recall, and F1 score. They all found that deep learning models perform well in classifying the phase of flight, demonstrating the value of these techniques for automating aviation safety analysis, with the large datasets of the NTSB and ATSB performing significantly better than the limited data set for the ASN.
1) [20] compares the performance of sRNN and ResNet, finding that sRNN outperformed the simplified ResNet, achieving 83% accuracy.
2) [21] focused on comparing the performance of four individual deep learning models, finding that LSTM and BLSTM achieved the highest performance with 87% accuracy.
3) [22] goes further by evaluating combinations of deep learning models, finding that combined models performed better, with the LSTM+BLSTM model achieving the highest accuracy at 67%.

These studies suggest that NLP and deep learning are promising for classifying the phase of flight from aviation safety reports, helping stakeholders analyze incidents more efficiently and accurately. Combining different deep learning models appears to improve classification accuracy, so future research should explore more complex model architectures. Additionally, incorporating features beyond textual narratives, such as flight data recorder information, could further enhance model accuracy and robustness. Again, the sample size appears to be a significant determinant of performance, requiring further investigation.

### C. Topic Modelling

Finally, we will compare the work on the use of TM for analyzing aviation safety reports, comparing performance across different sources [15, 24, 25]. All of these studies utilize NLP techniques with TM to analyze aviation incident or accident reports. The underlying goal was to uncover latent thematic structures within the textual data, leading to a deeper understanding of events, contributing factors, and potential areas for safety improvements in aviation. [15]. [24]. [25].

All three sources share a similar methodology, including data collection, text preprocessing, topic modeling, and model performance evaluation:
- They acquire data from publicly available aviation safety reports.
- Standard text preprocessing steps include tokenization, lowercasing, removal of punctuation, stop words, URLs, and HTML tags, with lemmatization to reduce words to their base form.
- Two feature extraction techniques are used: TF-IDF and Word2Vec.
- Coherence scores and interpretability metrics evaluate the effectiveness of the TM techniques.
- Python and libraries ( NLTK, Gensim, and Scikit-Learn) were used for implementing TM techniques.

In terms of the TM techniques
1) [15] used LDA and NMF,
2) [24] used pLSA, LSA, LDA, and NMF, and
3) [25] also used pLSA, LSA, LDA, and NMF, with K-means clustering.

The three studies differed in datasets, number of topics extracted, and specific keywords associated with the topics:
1) [15] & [25] used 36,000 records from the NTSB (2000-2020) and extracted 10 topics, while
2) [24] used 50,778 records from the ATSB (2013-2022) and extracts 10 topics.

Collectively, these studies highlight that different TM techniques offer varying strengths and weaknesses for uncovering latent thematic structures within textual data. Selecting the most appropriate TM technique depends on the specific objectives of the analysis and dataset characteristics. Figure 2 shows the associated word clouds from the TM utilizing the NMF model for the NTSB data (left) and ATSB data (right). An unimportant difference is the use of the term airplane in the US (NTSB) context, as opposed to aircraft in the Australian (ATSB) context. Approach is a similar size in both, with runway showing a significant difference in size. Further comparative analysis will be completed.

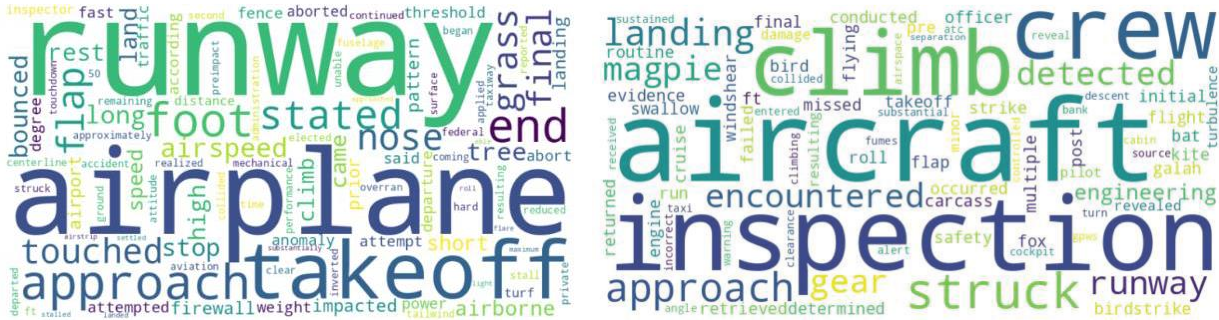

**Fig. 2  Word clouds for NMF model using the NTSB (left) and ATSB (right) data sets.**

## V. Quantitative Analysis

### A. Damage Level Identification

As noted previously, all of the models performed well, with combined models showing a slight advantage, but not enough to broadly apply then. Table 1 shows the model performance in terms of precision, recall, F1-score, and accuracy, for the best performing single model for both the NTSB and ATSB data sets. While the corpus was larger for the NTSB data set with narratives expected to have lengths of 36.6 words, compared to the ATSB with narratives expected to have lengths of 23.3 words, the performance is comparable. This is likely due to the shorter narratives of the ATSB still providing clear information such that the damage level can be accurately identified 89% of the time, relative to the 90% accuracy for the NTSB. Future work will look to use the ASN data set to provide a comparison, although the NTSB and ATSB both use four levels of damage (destroyed, substantial, minor, and none), while the ASN identifies accidents as A1 or A2, being either written off or repaired. Also, the expected narrative length of the ASN is 188.8 words. The other point to note there is that the NTSB and ATSB data sets include both accidents and incidents, while the ASN only includes accidents. As such, the relative comparison to the ASN in future work needs to consider these features.

**Table 1  Model performance for damage level identification.**

| Data | Model | Precision | Recall | F1-Score | Accuracy |
|---|---|---|---|---|---|
| **NTSB [8]** | sRNN | 82 | 90 | 86 | 90 |
| **ATSB [19]** | sRNN | 87 | 89 | 87 | 89 |

### B. Flight Phase Identification

All three data sets have been utilized to investigate model performance for flight phase identification. While combined models were considered when looking at the ASN data (in order to try and improve performance), the results in Table 2 compare single model performance metrics. Again, the model performance is given in terms of precision, recall, F1-score, and accuracy. In contrast to the damage level identification, the ATSB data clearly outperformed the NTSB data, but only by 3 to 5 percentage points. Also, the ATSB model that performed the best is now the LSTM, whereas for the damage level identification, both of the best performing models were the sRNN. It should be noted that there are now significantly more categories that need to be considered (seven phases of flight).

**Table 2  Model performance for flight phase identification.**

| Data | Model | Precision | Recall | F1-Score | Accuracy |
|---|---|---|---|---|---|
| **NTSB [20]** | sRNN | 84 | 83 | 83 | 83.2 |
| **ATSB [21]** | LSTM | 88 | 87 | 88 | 87.4 |
| **ASN [22]** | BLSTM | 63 | 64 | 63 | 64 |

For the much smaller dataset of the ASN, only 9% of the number of safety occurrences relative to the ATSB, the total corpus is also smaller (705k, compared to the 1M for the ATSB and 2.9M for the NTSB). This suggests that the number of narratives, with suitable information, is clearly more important to performance than the overall length of the narratives. This is shown by the 20-percentage point reduction in the performance across the board of the ASN model relative to the ATSB and NTSB models for flight phase identification.

Importantly here, all three data sets were used to classify the same sever phases of flight: approach, enroute, landing, standing, takeoff, taxi, and unknown. Hence while the reports are on average longer for the ASN, the amount of text that is needed to describe the phase of flight is clearly conveyed in the shorter narratives of the NTSB and the ATSB.

**C. Topic Modelling**

Figure 3 shows the associated topic distributions utilizing the LDA model for the NTSB data (left) and ATSB data (right). An unimportant difference is the use of the term airplane in the US (NTSB) context, as opposed to aircraft in the Australian (ATSB) context. Approach is a similar size in both, with runway showing a significant difference in size. The greatest coherence score for the TM of the NTSB data was 0.597 using LDA, which compares to the coherence score of 0.58 for the ATSB data, also using LDA.

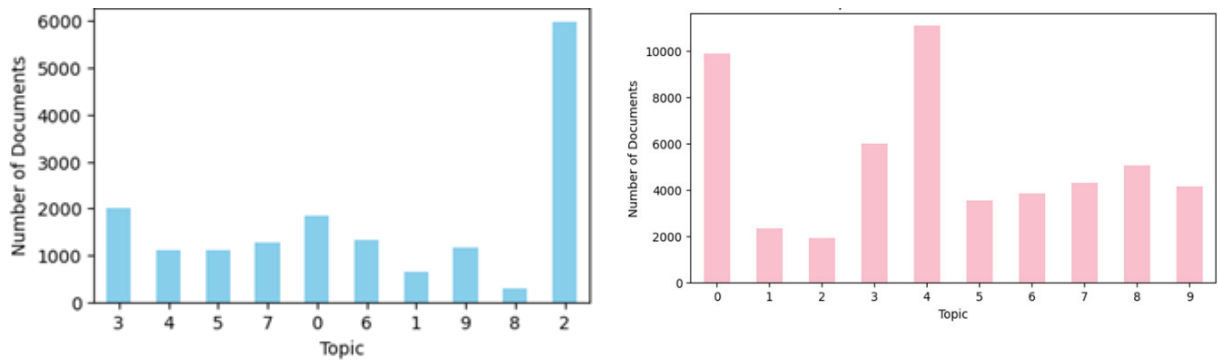

**Fig. 3  Topic distributions using LDA model for the NTSB (left) and ATSB (right) data sets.**

## VI.   Discussion

In recent years, the application of natural language processing (NLP) and machine learning to aviation safety analysis has gained significant attention [26]. Our research, explores various aspects of this field, including topic modeling, deep learning for categorization, and the impact of different datasets on model performance. This research and similar work conducted by others will provide a comprehensive understanding of the current state and future possibilities of these technologies in enhancing aviation safety.

We have investigated and compared different topic modeling methods, specifically pLSA, LSA, LDA, and NMF [15, 24, 25]. These studies highlight the strengths and limitations of each method. Notably, we focused on the NTSB dataset and demonstrated that LDA performs best with a coherence score of 0.597 [25]. This finding aligns with Paradis et al. [27], which suggests that LDA effectively uncovers hidden topics and trends in aviation safety data. Conversely, using the ATSB dataset, we reached a similar conclusion about the relative strengths of LDA and NMF but noted a smaller performance difference between the two [24]. This suggests that the effectiveness of different topic modeling techniques can vary based on the dataset used.

We have also focused on applying deep learning models for classifying aviation safety events. In our studies, we explored the classification of damage levels based on textual narratives from NTSB reports [8]. Our findings indicate that deep learning models, such as RNNs, LSTMs, GRUs, and their combinations, can achieve high accuracy in this task. These results support the idea presented by Zhang et al. [28] that sequential deep learning models like LSTMs, when used with word embedding techniques, have significant potential for predicting aviation safety outcomes. Additionally, we examined the categorization of the phase of flight during which an incident occurred, further highlighting the effectiveness of deep learning models [20-22]. While not directly addressed by other, the general features of the BERT for aviation demonstrate the ability to capture these features as well [11].

We examined how the size and source of datasets impact the accuracy of aviation safety incident prediction models [23]. Our findings suggest that larger datasets generally lead to higher accuracy, though valuable insights can still be drawn from smaller datasets. This observation is consistent with the broader discussion on the data requirements for deep learning models. While large datasets are typically deemed essential for optimal performance [29], our findings emphasize the need to explore techniques that can efficiently utilize varying datasets, particularly in fields like aviation safety.

A significant aspect of our research is the application of BERT and its variations for aviation safety analysis. This approach aligns with the broader trend in NLP, where transformer-based language models like BERT have gained prominence. Articles by Jing et al. [11], Chandra et al. [9], and Andrade and Walsh [10] provide further insights into the use of BERT for aviation-specific tasks. Jing et al. [11] explores the use of BERT for aviation text classification, highlighting its superior performance compared to traditional methods. Chandra et al. [9] proposes an aviation-specific BERT model, named Aviation-BERT, which has been pre-trained on aviation-related text data and demonstrates improved performance in tasks like masked word prediction. Furthermore, Andrade and Walsh [10] introduce SafeAeroBERT, a safety-informed aerospace-specific language model based on BERT. These developments underscore the growing interest and success in adapting BERT for aviation applications.

Our ongoing research, when viewed in the broader context of the field, offers insights for both researchers and practitioners in the aviation safety industry. Our analysis of different NLP and ML techniques, coupled with our investigation of various datasets, shows the potential and limitations of these methods. This comparative viewpoint emphasizes the importance of carefully evaluating both the technical elements of model selection and the characteristics of the available data to achieve meaningful and impactful results that advance the analysis of aviation safety.

## VII. Conclusion

This work has explored the use of AI, particularly NLP and deep learning techniques, for analyzing aviation safety reports, classifying damage level, identifying flight phase during incidents, and uncovering thematic structures through TM. The research presented suggests that these techniques have the potential to revolutionize aviation safety analysis. The comparative analysis of various deep learning models and TM techniques applied to datasets from the NTSB and ATSB highlights the impact of dataset size and source on the accuracy of the analysis. The findings consistently demonstrate that larger datasets, like those available from the NTSB, generally lead to higher accuracy in both damage level and flight phase classification tasks. However, the study also reveals that even with smaller datasets, valuable insights can be obtained. Specifically, the application of TM to both NTSB and ATSB datasets has been successful in uncovering recurring themes and patterns, such as mechanical failures and fuel system issues, regardless of the difference in dataset size. This underscores the importance of choosing the appropriate technique based on the specific research goals and dataset characteristics. The insights gleaned from this research can inform the development of more proactive safety management systems, helping to prevent future incidents and enhance aviation safety. Future research will further explore the impact of dataset characteristics, such as narrative length and the inclusion of both accidents and incidents, on model performance. Additionally, the integration of other data sources, such as flight data recorder information, will be investigated to enhance model accuracy and robustness.